\documentclass[letterpaper, 10 pt, conference]{ieeeconf}  

\IEEEoverridecommandlockouts                              

\usepackage{graphicx}

\usepackage{todonotes}
\usepackage{algorithm}
\usepackage{algpseudocode}
\usepackage{amsmath,amsfonts}

\DeclareMathOperator*{\argmax}{argmax}

\title{\LARGE \bf
Inverse Reinforcement Learning for Strategy Identification
}

\author{Mark Rucker$^{1}$, Stephen Adams$^{2}$, Roy Hayes$^{3}$, and Peter A. Beling$^{4}$
\thanks{$^{1}$Mark Rucker is a Graduate Research Assistant in the Department of Engineering Systems and Environment at the University of Virginia.
        {\tt\small mr2an@virginia.edu}}%
\thanks{$^{2}$Stephen Adams is a Principal Scientist in the Department of Engineering Systems and Environment at the University of Virginia.
        {\tt\small sca2c@virginia.edu}}%
\thanks{$^{3}$Roy Hayes is the CTO of Systems Engineering, Inc. and graduated from the University of Virginia with a PhD. in Systems Engineering.
        {\tt\small rlh8t@virginia.edu}}%
\thanks{$^{4}$Peter A. Beling is a Professor in the Department of Engineering Systems and Environment at the University of Virginia.
        {\tt\small pb3a@virginia.edu}}%
}

\begin{document}

\maketitle
\thispagestyle{empty}
\pagestyle{empty}

\begin{abstract}

In adversarial environments, one side could gain an advantage by identifying the opponent's strategy.  For example, in combat games, if an opponents strategy is identified as overly aggressive, one could lay a trap that exploits the opponent's aggressive nature. However, an opponent's strategy is not always apparent and may need to be estimated from observations of their actions.  This paper proposes to use inverse reinforcement learning (IRL) to identify strategies in adversarial environments.  Specifically, the contributions of this work are 1) the demonstration of this concept on gaming combat data generated from three pre-defined strategies and 2) the framework for using IRL to achieve strategy identification.  The numerical experiments demonstrate that the recovered rewards can be identified using a variety of techniques.  In this paper, the recovered reward are visually displayed, clustered using unsupervised learning, and classified using a supervised learner.

\end{abstract}

\section{INTRODUCTION}

In adversarial environments, such as sporting events, gaming, or even defending against a cyber attack, one side could gain an advantage by identifying the opponent's strategy.  The value of strategy identification lies in the ability it provides to foresee, and thereby counter, the opponent's future actions.  In combat games, for example, if an opponent's strategy is identified as overly aggressive, one could lay a trap that exploits the opponent's aggressive nature. However, an opponent's strategy is not always apparent and may need to be estimated from observations of their actions.  This paper proposes to use machine learning, specifically inverse reinforcement learning, to identify strategies in adversarial environments.

Strategic planning is often defined by four concepts.  \emph{Goals} are high-level concepts that define what needs to be accomplished.  \emph{Objectives} are quantitative measures that determine if a goal has been achieved.  \emph{Strategies} are plans for achieving the defined objectives.  \emph{Tactics} are the low-level actions for carrying out a strategy.  In the combat gaming example, the goal is to win the engagement.  There are numerous objectives that could be associated with achieving this goal, including securing a particular target or minimizing casualties.  Similarly, there are many strategies for achieving objectives; for example, being aggressive or defensive.  Tactics are the specific sequence of actions carried out by each side in the engagement.  The high-level goal (to win) is known to both sides, and the low-level tactics are observable during the engagement.  However, the objectives and the strategies of the opponent are unknown, but given the hierarchy previously defined the objectives and the strategies should be correlated with the tactics.  Therefore, we hypothesis that the objectives and the strategies can be inferred from the observed tactics of the opponent.

Markov decision processes (MDPs) \cite{puterman2014markov,scherer2018practical} are often used in artificial systems to model sequential decision making.  An MDP is primarily defined by states, actions, rewards, and a state transition function.  A policy maps states to actions, and an optimal policy maximizes expected future reward.  The components of an MDP map to the four strategic planning concepts previously defined.  The \emph{goal} is to maximize expected future reward.  The \emph{tactics} are the state-actions pairs that result from an implemented policy.  The \emph{objectives} and the \emph{strategy} are represented by the reward function, i.e. the reward function is measurable and the parameters of the reward function define an agent's behavior. 

Reinforcement learning (RL) \cite{sutton2018reinforcement} is a common machine learning technique for learning optimal policies through interaction with an environment.  The standard RL paradigm involves an agent selecting an action in a given state and then observing a state transition defined by the transition function and a reward defined by the reward function.  By repeating this process and exploring using a probabilistic exploration policy that includes random actions, the agent can eventually learn an optimal policy.  On the other hand, inverse reinforcement learning (IRL) \cite{ng2000algorithms} estimates a reward function from observations of an expert's actions in an environment.

Inverse reinforcement learning has developed largely within the field of robotics. A few of the notable advances in robotics include training autonomous acrobatic helicopters \cite{abbeel2007application}, robot navigation through crowded rooms \cite{kretzschmar2016socially}, and autonomous vehicles \cite{you2019advanced}. More recently there has been growing interest in applying IRL to human behavior problems in areas such as inferring cultural values \cite{nouri2012cultural}, modelling human routines \cite{banovic2016modeling}, and even so far as neuroscience models of human reasoning \cite{jara-ettinger2019theory}. Unfortunately, theoretical solutions to the IRL problem remain challenging, requiring either strict mathematical assumptions such as linearity (e.g., \cite{abbeel2004apprenticeship,ziebart2008maximum}) or large resource requirements such as sample or computation complexity (e.g., \cite{qiao2011inverse,wulfmeier2017large}). As such, the field of IRL is a very active one that has huge potential as the theory continues to advance. 

Multi-agent IRL estimates reward functions in multi-agent settings and most techniques rely heavily on game-theoretic principles \cite{lin2017multiagent,lin2019multi}.  However, this reliance makes their use impractical when notions of equilibria cannot be defined.  At a high level, multi-agent IRL seems like a suitable construct for strategy identification in combat scenarios, e.g. each side could be considered an agent and each side may be composed of multiple agents, but the multi-agent IRL construct and algorithms need significant advancement before they can become feasible. 



This paper proposes that IRL can be used to recover unique reward functions that are correlated with strategies in adversarial environments such as combat games.  Specifically, the contributions of this work are 1) the demonstration of this concept on gaming combat data using three pre-defined strategies and 2) the framework for using IRL to achieve strategy identification.  The IRL algorithm used in this study utilizes kernel methods to realize expressive functions of the state space \cite{rucker2020human,kim2018imitation}.  IRL has been used to create agent based models \cite{lee2017agent}, identify decision agents \cite{qiao2013recognition}, and to estimate strategy in other domains, including animal behavior \cite{yamaguchi2018identification}, table tennis \cite{muelling2013inverse}, and financial trading \cite{yang2015gaussian,yang2014algorithmic}. However, this is the first study to apply IRL to estimating strategies for combat games.  

Further, IRL has two traits that make the technique well suited for strategy identification.  First, the reward function is independent of the environment dynamics, but the policy is dependent upon the both the reward function and the environment dynamics.  This is analogous to the concepts in strategic planning where the strategy and objectives are most likely independent of low-level concepts like setting but the tactics are not.  Second, the estimated reward function can be used for prediction if the dynamics change.  Simply estimating the policy in this situation is not sufficient as it may no longer be optimal and therefore may not be directing the agent's actions.  The reward function could be used in conjunction with a model-free learner to estimate a new policy and thus predict future actions of the adversary.        

This paper is organized as follows. Section \ref{sec:background} provides background information on MDPs, RL, and IRL.  Section \ref{sec:alg} outlines the IRL algorithms used for strategy identification.  Section \ref{sec:exp} describes the numerical experiments performed to validate the IRL methods.  Section \ref{sec:concl} provides our conclusions from the numerical experiments and possible areas of future work.  

\section{BACKGROUND}\label{sec:background}

General RL theory rests on Markov decision process models (MDPs). In this paper an MDP is defined as any tuple $(S, A, \mathcal{T}, r)$ where $S$ is a set of states, $A$ is a set of actions, $\mathcal{T}: S \times S \times A \to [0,1]$ is a transition probability function satisfying $\mathcal{T}(s',s,a) = P(s'\mid s,a)$, and $r: S \to \mathbb{R}$ is a reward function. 

Given an MDP any function $\pi: A \times S \to [0,1]$ satisfying $\pi(a,s) = P(a \mid s)$ will be called a policy function. Every $\pi$ possesses a unique value function $V^{\pi}: S \to \mathbb{R}$ and action-value function $Q^{\pi}: S \times A \to \mathbb{R}$ defined as
\begin{align*}
    V^{\pi}(s) &= \lim_{T\to\infty} \mathbb{E}_{\pi}\left[\sum_{t=1}^{T} \gamma^{t-1} r(X_t) \Big| X_1 = s \right] \\
    Q^{\pi}(s,a) &= \lim_{T\to\infty} \mathbb{E}_{\pi}\left[\sum_{t=1}^{T} \gamma^{t-1} r(X_t) \Big| X_1 = s, Y_1 = a \right].
\end{align*}
where $\gamma \in [0,1)$ is called a discount factor, $X_t(\omega) = s_t \in S$ is a random variable defined on $\omega = (s1,a1,s2,a2,\ldots) \in \Omega = (S \times A)^\infty$ and $\omega$ is distributed according to 
\begin{equation}
P_{\pi}(\omega \mid s_1) = \prod_{t=1}^{\infty} \pi(a_t,s_t) \mathcal{T}(s_{t+1},s_t,a_t).\label{eq:2}
\end{equation}

The goal of RL algorithms is to learn an optimal policy, $\pi^*$, for an MDP where optimality is defined as $V^{\pi^*}(s) \geq V^\pi(s)\ \forall \pi, s$. The goal of IRL algorithms is to learn a reward $r$, assuming an MDP where $r$ is not known, for which a given policy $\pi$ is optimal.  Because the traditional IRL problem statement has known degenerate solutions additional requirements are often added to yield useful solutions.

\begin{table*}
\caption{This table provides the summary statistics for 36 simulated engagements. The left column of the table indicates the AI strategy for the blue force in the engagement. The tSNE plot later in this paper is visualizing the reward functions learned from the blue force in these matches.}\label{tab2}
\centering
\begin{tabular}{|c||c|c|c|c|c|c|c|c|c|c|c|c|}

\hline
 & Match & Red & Blue & \multicolumn{3}{c|}{Match Time} & \multicolumn{3}{c|}{Red Deaths} & \multicolumn{3}{c|}{Blue Deaths}  \\
 \hline
 
 & Count & Count & Count & Min & Avg & Max & Min & Avg & Max & Min & Avg & Max  \\

\hline
\hline

Fallback & 11 & 12 & 11 & 54s & 202s & 468s & 1 & 7.6 & 11 & 1 & 7.1 & 11 \\
\hline

Assault & 12 & 12 & 11 & 57s & 192s & 402s & 4 & 9.3 & 11 & 3 & 8.6 & 11 \\
\hline

Flank & 13 & 12 & 11 & 54s & 162s & 279s & 1 & 6.8 & 11 & 3 & 8.1 & 11 \\
\hline

\end{tabular}
\end{table*}

\begin{table}
\caption{The full definition of the MDP.}\label{tab1}
\begin{tabular}{|p{.5cm}|p{7cm}|}

\hline
Item & Definition \\

\hline
$S$ & A state is defined as the current match time, the x,y position and health of all red and blue forces (sans one blue force player), and the position and health of our agent (who replaces the missing blue force player). \\

\hline 
$A$ & There are 9 actions: stay in place or move one step in one of the 8 cardinal/ordinal directions. \\

\hline 
$\mathcal{T}$ & All transitions are deterministic with our controlled agent moving according to the action selection and all other agents moving to their positions according to the pre-recorded data  \\

\hline
$r$ & Unknown \\

\hline
\end{tabular}
\end{table}

\section{ALGORITHMS}\label{sec:alg}

\subsection{Reward Learning}

\subsubsection{IRL Algorithm}
To learn rewards the projection variant of \cite{abbeel2004apprenticeship} was utilized and extended via kernel-based methods as described in \cite{rucker2020human}. For our purposes we define a kernel as any $k: S \times S \to \mathbb{R}$ that induces a positive semi-definite Gram matrix $K \in \mathbb{R}^{S \times S}$ where $K_{ij}(S,S) = k(S_i, S_j)$.

Intuitively, kernel methods can be thought of as either a non-linear extension to linear function approximation techniques or as functions in a space $H = \operatorname{span}(k(S,\cdot))$. A particularly common interpretation for kernels is as similarity measures. This interpretation is most often taken when $k$'s value is in $[0,1]$.

Mathematically, kernels can be interpreted as an inner product between vectors in $H$. That is, $\langle s, s' \rangle_k = \langle k(s,\cdot), k(s',\cdot) \rangle := k(s,s')$. This definition also allows for the norm $|| v \in H|| := \sqrt{\langle v, v \rangle} $ which is useful for cluster analysis and visualizations.

In the original projection IRL algorithm, which is extended below, the goal is to match feature expectation given some $\Phi: S \to \mathbb{R}^n$. In the kernel-based extension feature expectation becomes the kernel expectation, defined as

\begin{equation} 
    \mu(\pi) = \mathbb{E}_{\pi}\left[\sum_{t=1}^T \gamma^{t-1}k(X_t, \cdot )\right]. \label{eq:1}
\end{equation}

The complete kernel-based algorithm is provided in Algorithm \ref{alg:IL}. It should be noted that the algorithm as stated creates a set of reward functions whose optimal policies have a convex combination that approximates the expert. For our analysis we required a single reward function and simply chose the $r_i$ from this set whose optimal policy $\pi^*_i$ had the smallest $||\mu_E-\mu(\pi^*_i)||$.

\begin{algorithm}
    \caption{Kernel-based Projection IRL}\label{alg:IL}
    \begin{algorithmic}[1]
        \State \textbf{Initialize:} set $\epsilon$ as desired
        \State \textbf{Initialize:} set $i \gets 1$ and $\mu_E \gets \mu(\pi_E)$
        \State \textbf{Initialize:} set $r_1$ to a random reward
        \State \textbf{Initialize:} set $\pi^*_1 \gets \pi^*$ for $r_1$ and $\bar{\mu}_1 \gets \mu(\pi_1^*)$
        \item[]
        \While  {$||\mu_E-\bar{\mu}_i|| > \epsilon$}
            \State $i \gets i+1$
            \State $\alpha_i \gets \mu_E - \bar{\mu}_{(i-1)}$
            \State $R_i(s) \gets \langle \alpha_i, k(s,\cdot) \rangle$
            \State $\pi^*_i \gets \pi^*$ for $r_i$ \label{ln:8} \Comment{see Algorithm \ref{alg:FL}}
            \State $\mu_i \gets \mu(\pi^*_i)$
            \State $\beta_i \gets \frac{ \langle \mu_i - \bar{\mu}_{(i-1)},\ \mu_E - \bar{\mu}_{(i-1)} \rangle}{ \langle \mu_i - \bar{\mu}_{(i-1)},\ \mu_i - \bar{\mu}_{(i-1)} \rangle}$
            \State $\bar{\mu}_i \gets \bar{\mu}_{(i-1)} + \beta_i\left(\mu_i - \bar{\mu}_{(i-1)}\right)$
        \EndWhile
    \end{algorithmic}
\end{algorithm}

\subsubsection{IRL Implementation}

Our implementation of the above algorithm uses an empirical estimate of $\mu_E$ calculated from some sample $E$ of observed expert trajectories  $(s_1, a_1, s_2, a_2, \ldots)$ drawn with probability $P(\pi_E \mid s_1)$ as defined in Equation \ref{eq:2}. Several estimation techniques were tested and the one that seemed to produce the best rewards (determined via human inspection) was

\begin{equation}
    \mu(\pi_E) \approx \hat{\mu}_E = \frac{T}{N(E)} \sum_{\omega}^{E} \sum_s^{\omega} k(s,\cdot)\label{eq:3},
\end{equation}

\noindent where $T$ was chosen arbitrarily to be 20 and $N$ is a function which returns the total number of states in $E$. Such a formulation means that longer episodes have more weight since they'll have more states and $T$ determines how far into the future to consider when estimating the expert's reward. Equation \ref{eq:3} was also used when estimating $\mu_i$ on line 4 and 10 in Algorithm \ref{alg:IL}.

The final component that needs to be implemented for a specific algorithm is the kernel $k$. For our kernel we calculated 6 features to describe the present location of our agent: (1) min distance to living red, (2) max distance to living red, (3) min distance to living blue, (4) max distance to living blue, (5) min cos similarity between living red and blue and (6) max cos similarity between living red and blue. These features were scaled appropriately so that each feature was in $[0,1]$ and then a Gaussian kernel was applied.

\subsection{Policy Learning}
\subsubsection{RL Algorithm}

\begin{figure*}
    \centering
    \includegraphics[width=\textwidth]{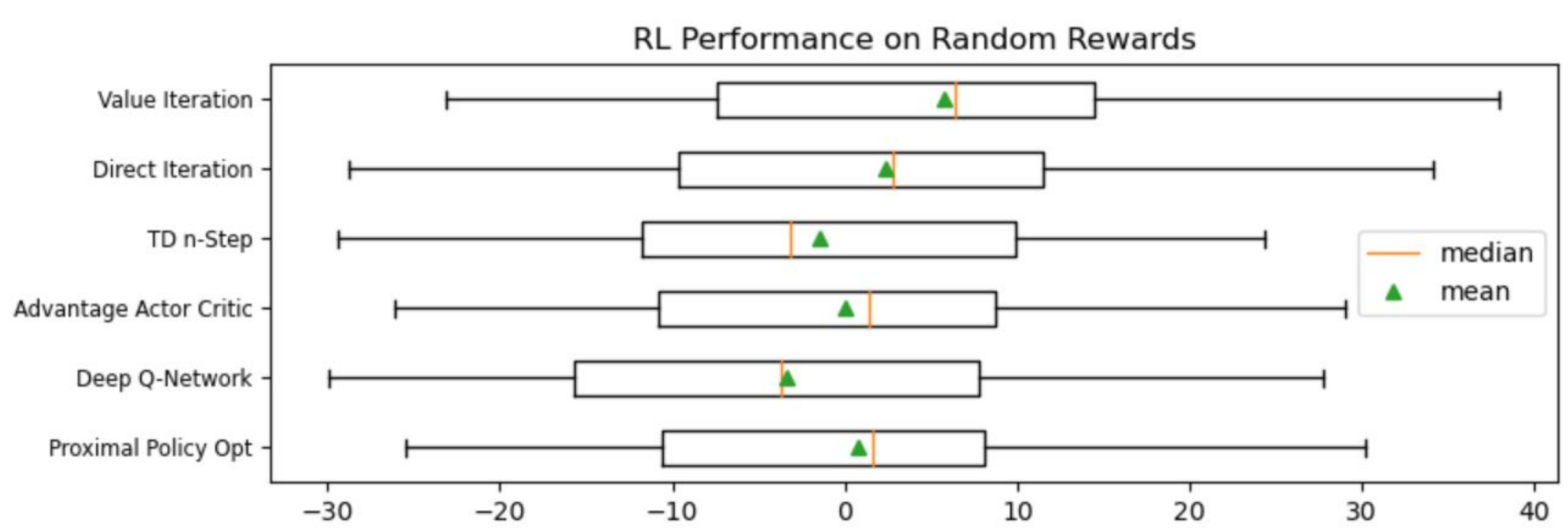}
    \caption{This figure shows the distribution of expected value given our MDP and 30 random kernel-based reward functions using the same kernel we use in KPIRL. Our method, direct iteration, out-performs other well-known RL algorithms though still falls short of the optimal value iteration. We believe our improved performance is largely due to the simplicity of our MDP and small sample sizes when training rather than an algorithmic advancement, though more analysis is necessary to confirm.} \label{fig:RL_perf}
\end{figure*}

The utilized IRL algorithm requires solving for optimal polices given a reward function on each iteration. To satisfy this requirement, an empirical estimate policy iteration method was used. Our approach is an n-step, model-free, Monte Carlo, on-policy Q-function approximation (see Algorithm \ref{alg:FL}) and was originally described in \cite{rucker2020human}.

Experiments show that our direct iteration method out-performs other well-known RL algorithms (see Figure \ref{fig:RL_perf}) given relatively small training samples from our MDP. All comparison RL algorithms come from the Stable Baseline 3 project \cite{stable-baselines3} and have been moderately tuned with respect to hyper-parameters. Every algorithm was given a budget of 10,000 interactions with the environment before comparing the results.

We believe our improved performance is largely due to the simplicity of our MDP and small sample sizes when training rather than an algorithmic advancement, though more analysis is necessary to confirm. For reference purposes Figure \ref{fig:RL_perf} also includes value iteration which represents the optimal solution. During final analysis value iteration was not used due to intractable memory on the full problem.

\begin{algorithm}
    \caption{Direct Estimate Iteration} \label{alg:FL}
    
    \begin{algorithmic}[1]
        \State \textbf{Initialize:} set $I$, $M$, $T$ and $W$ for iterations, episodes, steps and obs/episode
        \State \textbf{Initialize:} set $\pi_1$ to random policy
        \State \textbf{Initialize:} set $O \subseteq S \times A \times \mathbb{R} = \emptyset$

        \item[]
        \For {$i \gets 1$ to $I$}
            
            \item[]
            \For {$m \gets 1$ to $M$}
            
                \item[]
                \State Generate $s_0$ according to some distribution
                \State Generate $a_0$ according to some policy
                \State Given $s_0,a_0$ generate $s_1,a_1,\ldots,s_T$ with $\pi_i$

                \item[]
                \For {$w \gets 0$ to $T-W$}
                    \State $v \gets \sum_{t=0}^{W-1} \gamma^{t-1}r(s_{w+t})$
                    \If { $(s_w, a_w) \in O$ } 
                        \State $O_{(s_w, a_w)} \gets (O_{(s_w, a_w)} + \hat{v})/2  $
                    \Else
                        \State $O_{(s_w, a_w)} \gets \hat{v}$
                    \EndIf
                \EndFor
                
            \item[]
            \EndFor
            
            \item[]
            \State Fit a regressor $Q_i$ using $O$
            \State $\pi_{i+1}(a,s) \gets \textbf{1}_{\{a = \argmax_{a'} Q_i(s,a')\}}$
        
        \item[]
        \EndFor
    
    \item[]
    \State \textbf{Return:} $\pi_{I+1}$
    
    \end{algorithmic}
\end{algorithm}

\subsubsection{RL Implementation}
When implementing Algorithm \ref{alg:FL} hyper parameter tuning was used to select the values for $I$, $M$, $T$ and $W$ which gave the best expected value for random rewards. Of the four parameters, only $W$ resulted in counter-intuitive behavior, demonstrating performance degradation if $W$ was too small or too large. For $I$, $M$ and $T$ performance generally increased monotonically with decreasing returns.

Additionally, separate experiments were conducted to evaluate various regression learners for line 13 in Algorithm $\ref{alg:FL}$. We tested an SVM kernel regressor, a linear regressor, an AdaBoost regressor and a decision tree regressor. Of these regressors the decision tree performed best when using the same features as those used to calculate the IRL reward kernel.

Line 7 in Algorithm \ref{alg:FL} allowed for exploring starts. Four exploration heuristics were evaluated: (1) random selection, (2) greedy selection, (3) epsilon-greedy selection, and (4) softmax selection. Of these four methods for exploring starts softmax selections performed best.

Finally, three other variations on the algorithm were tested. First, bootstrapping the update target resulted in decreased performance. Second, using an every visit MC target (cf. \cite{singh1996reinforcement}) for updates instead of an n-step target gave a small increase in performance for small state spaces but a large decrease in performance in large state spaces. And third, modifying the algorithm to only use value observations from the most recent policy iteration (i.e., clearing $O$ between line 8 and 9 in Algorithm \ref{alg:FL}) resulted in decreased performance across the board.

\section{NUMERICAL EXPERIMENTS}\label{sec:exp}

\begin{figure}
\centering
\includegraphics[width=6cm]{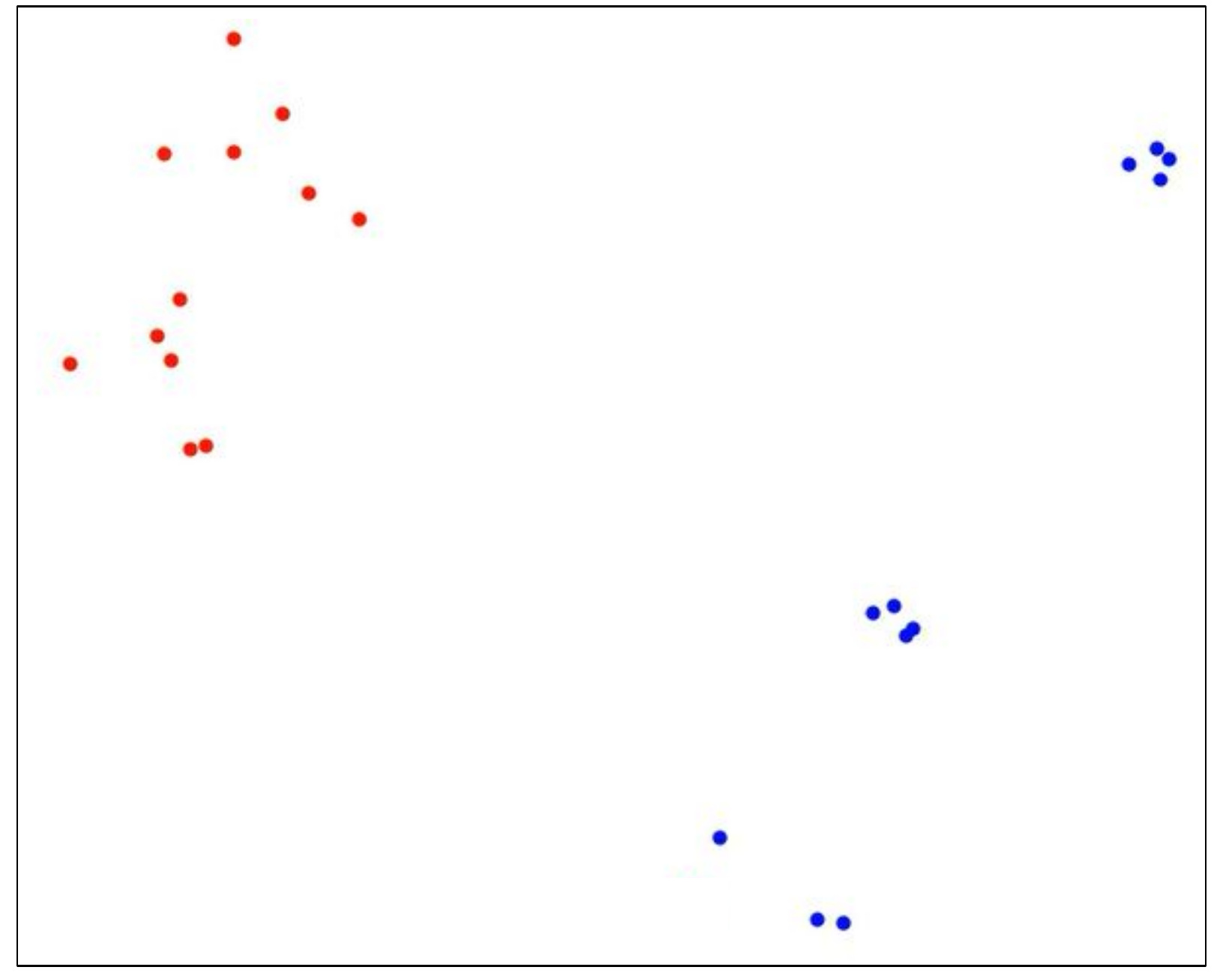}
\caption{The initial starting positions of the red and blue forces in one of the 36 generated matches. These engagements occurred on flat, open terrain inside an area of 340 by 340 meters. } \label{fig:example_states}
\end{figure}

\begin{figure*}
    \centering
    \includegraphics[scale=.95]{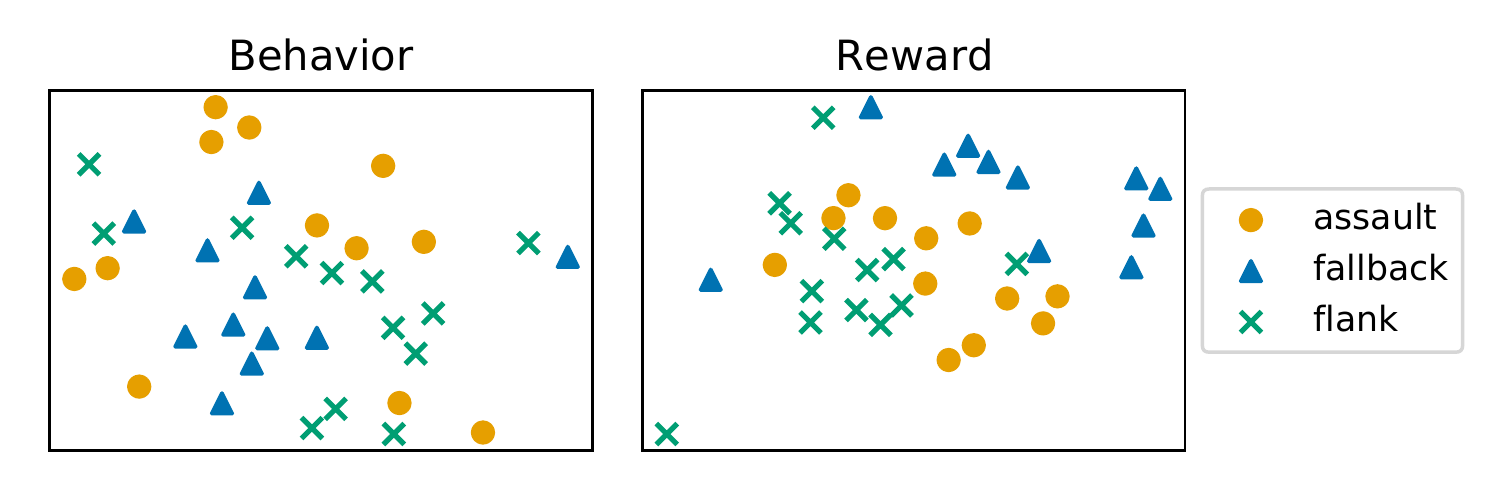}
    \caption{Two t-SNE plots were generated. The left represents expert kernel expectations (cf. Equation \ref{eq:1}) which we call behavior due to it being directly calculated form observed behavior. The right represents reward functions learned from the expert kernel expectations on the left. } \label{fig3}
\end{figure*}

Using the above algorithms three experiments were performed to explore the strengths and weaknesses of using IRL to analyze strategic behavior. To drive these experiments 36 matches involving combat engagements between two opposing forces (referred to as red and blue) were simulated using a gaming combat simulator. In each match both forces were comprised of three fireteams containing three to four AI controlled players each. Matches always occurred in an open field, and the amount of initial space separating the forces was varied slightly (see Figure \ref{fig:example_states} for an example starting position in an match). At a low level players were controlled by the native AI within the combat game. At a high level forces were nudged to follow one of three strategies: assault, flank or fallback. The red force always assaulted while the blue force assaulted 12 times, flanked 13 times and fellback 11 times. Summary statistics for the data can be found in Table \ref{tab2}.

During the 36 simulated matches the location of every AI player, which side they were on, and their health was recorded every 3 seconds. These observations were stored and used during IRL analysis to construct expert state trajectories. Because the IRL algorithm used kernel methods we needed to define $k: S \times S \to \mathbb{R}$. In this analysis, $k$ was a similarity kernel that mapped pairs of states to real numbers between 0 and 1 with pairs of similar states mapping closer to 1 and pairs of dissimilar states mapping closer to 0. When determining the similarity of two states $k$ considered the minimum distances to blue and red forces, the maximum distances to blue and red forces, and the angles of fire between blue and red forces.

In addition, to learn rewards from the above data using the algorithms in this paper the forward RL problem needed to be solved. To facilitate this a simplified MDP simulation was developed where we could take over individual agents and move them at will through any of the 36 pre-recorded matches. Our MDP was far simpler than the original simulation in that it only generated data at the same fidelity of the recorded observations and the environment did not respond differently to deviations in our controlled agent's behavior.

Despite these huge simplifications IRL was still able to learn meaningful rewards. The full description of the states, actions and transition function in the simulated MDP environment are provided in Table \ref{tab1}.

Using the 36 recorded data sets from the combat game, the kernel $k$, and the simplified MDP it was then possible to both empirically estimate the kernel expectation for the expert policy (see Equation \ref{eq:1}) in each data set as well as well as learn a reward function $R \in \operatorname{span}(k(S,\cdot))$ which generated a policy similar to the expert. Using these component, a labeled test was constructed for strategy identification experiments. The label was the blue force's high level strategy directive, behavior was represented as the empirical kernel expectation for the blue force, and reward was represented as the kernel-based reward function learned via IRL for the blue force.

Three experiments were conducted using the strategy labeled data set: (1) see if a t-SNE plot would show visible separation of strategies with either the expert kernel expectations or reward functions, (2) see if unsupervised learning techniques would be able to cluster strategies with either expert kernel expectations or reward functions and (3) see if a classifier could be trained to identify strategy given either an expert kernel expectation or reward function.

To conduct the t-SNE experiment it was necessary to calculate a distance matrix for both the kernel expectations and the reward functions. This was done via the norm induced by $k$. For example, if $a = k(s_1,\cdot) + 2k(s_2, \cdot)$ and $b = k(s_2, \cdot)$ then the distance between $a$ and $b$ was calculated as 
\begin{align*}
    ||a-b||^2 &= \langle a-b, a-b \rangle \\
    &= \langle k(s_1, \cdot) + k(s_2, \cdot), k(s_1, \cdot) + k(s_2, \cdot) \rangle \\
    &= k(s_1,s_1)+k(s_2,s_2)+2k(s_1,s_2).
\end{align*}
The t-SNE plot generated from the distance matrices seemed to show slight visual improvements in strategy separation for reward functions over their subsequent expert kernel expectation (see Figure \ref{fig3}). 

While care should be taken when interpreting t-SNE plots it was interesting to see that when visualizing reward functions flanking strategies were placed in between assault and fallback strategies. This aligned with the data recordings where assaults tended to be direct charges, fallback tended to be direct retreats and flanking strategies tended to be a mix of strategic approaches and fallbacks.

\begin{figure*}
    \centering
    \includegraphics[width=\textwidth]{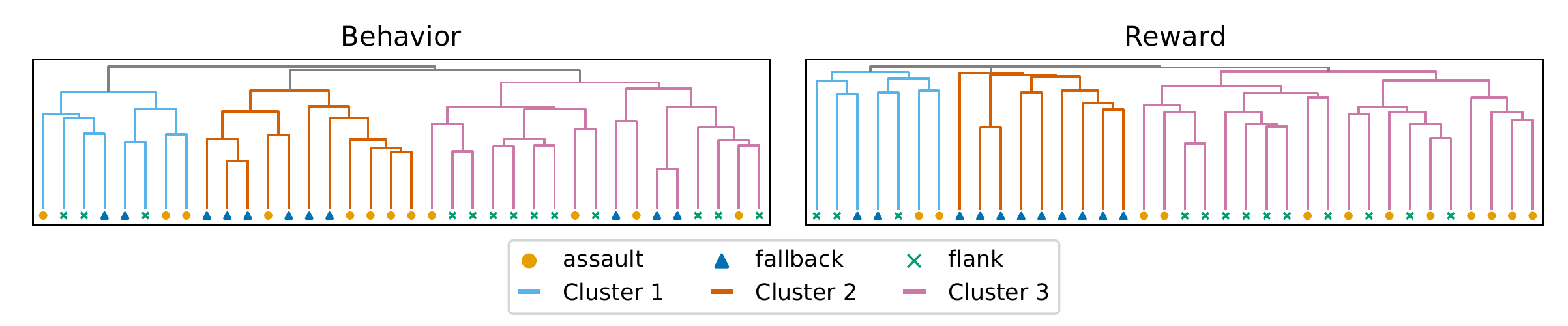}
    \caption{Two dendrograms were generated from the expert kernel expectations and their learned reward functions using hierarchical agglomerative clustering. The colors in each plot represent best separation for 3 clusters (the known right answer). The clusters in the behavior dendrogram show relatively little pattern. In the reward dendrogram on the other hand two clear clusters emerge, a fallback cluster and an  assault/flanking cluster. The third cluster shows little pattern. These results align with those seen in the t-SNE plots. } \label{fig4}
\end{figure*}

To conduct the cluster experiments square distance matrices were needed once again. As before these were calculated using the norm induced by $k$. Using the distance matrices, a hierarchical agglomerative clustering algorithm, and a complete linkage function two dendrograms were generated (see Figure \ref{fig4}). Clear patterns can be seen in the reward function clusters where 82\% of fallback strategies belong to the second cluster and 80\% of flank/assault strategies belong to the third cluster.

An interesting pattern that emerged in the cluster analysis was the first cluster in the kernel-based reward functions. At first glance there is no apparent pattern to this cluster which contains 3 flank, 2 fallback and 2 assault data sets. However, upon further analysis it was found that these 7 data sets represented all the matches where red and blue forces were placed within very close proximity before starting the match. Matches that began with forces in close proximity were considerably more chaotic than mid range and far matches. In close matches units often died quickly and the underlying combat simulator AI had little time to plan for the strategy nudges that were provided.

\begin{figure*}
    \centering
    \includegraphics[scale=0.8]{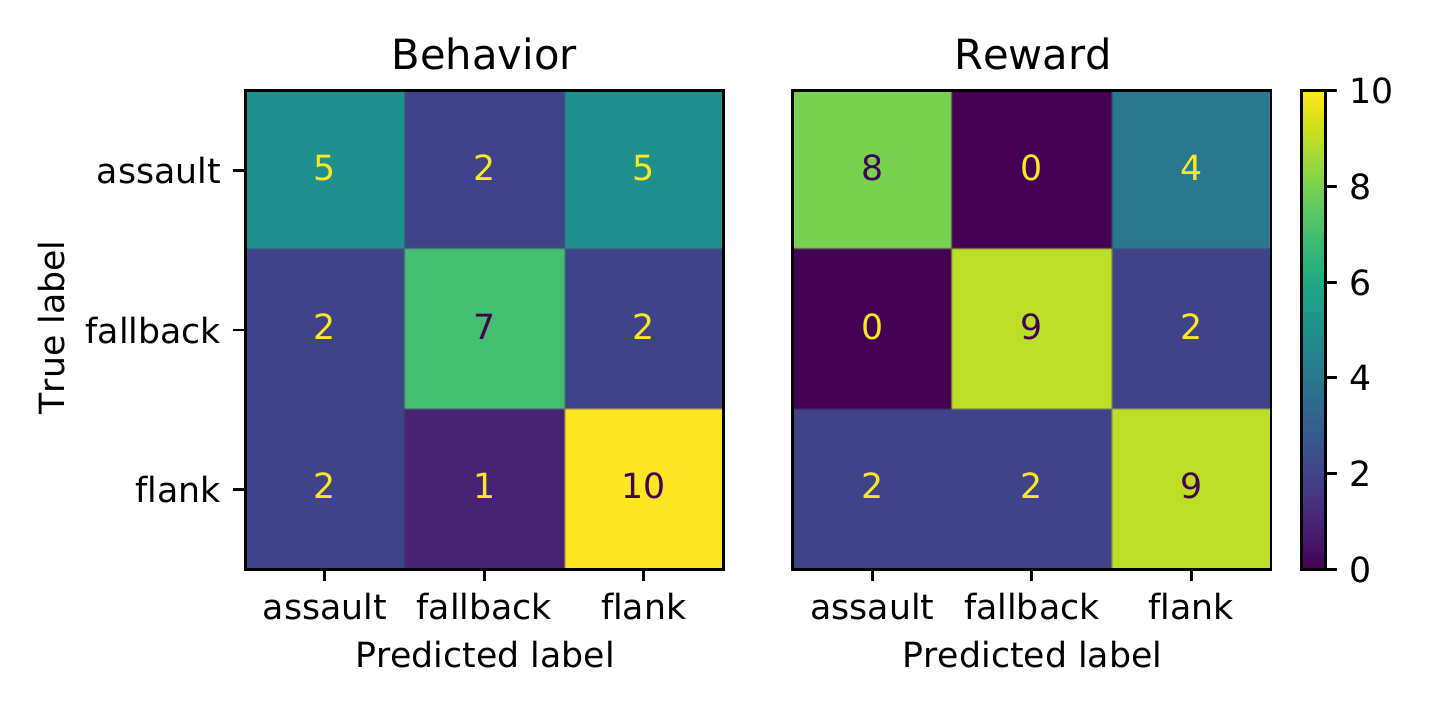}
    \caption{Two confusion matrices were created: one from a classifier which predicts strategy from expert kernel expectations and one from a classifier which predicts strategy from the learned reward functions. The reward classifier dominated kernel expectation classifier in all classes except flank. } \label{fig5}
\end{figure*}

For the final experiment a kernel-based SVM classifier was trained using a one-to-one approach to handle three classes. The classifier was evaluated on overall prediction accuracy using leave one out on all training points. Using expert kernel expectation gave an overall accuracy of 61\% while reward functions gave an overall accuracy of 72\%. The class specific breakdowns can be seen in the confusion matrix in Figure \ref{fig5}.

The results here followed a similar pattern to the two previous experiments. The classifier from the rewards outperformed the classifier from the kernel expectations, and assault and fallback strategies were easiest to distinguish while flanking and assaulting were commonly mislabeled. The four reward classifier mistakes that confused flanking and fallback data also belonged to the first cluster in the reward dendrogram. Visual inspection found that these four data sets had many early deaths making it difficult for the low level AI to form coherent strategies.

\begin{figure}
    \centering
    \includegraphics[width=8cm]{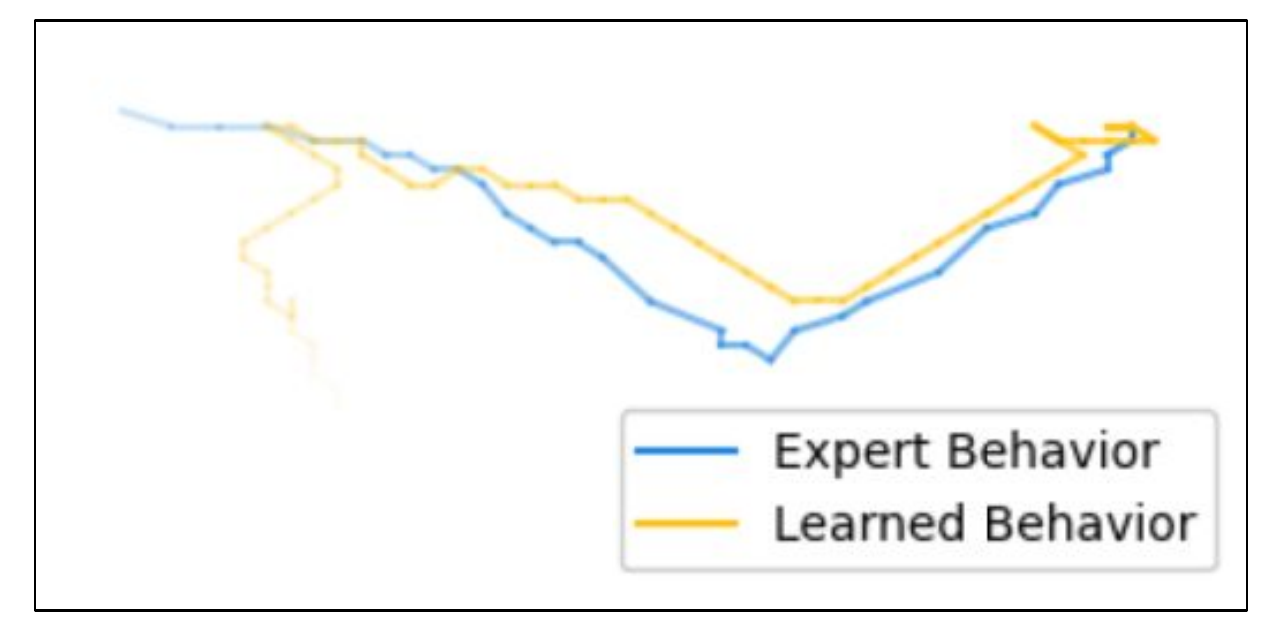}
    \caption{Shown above is a direct overlay of an expert's behavior with a policy learned from an IRL reward function. The gradient in the lines represents time with the darker portions of the line being earlier in time and the lighter portions of the line later in time. At the end of the trajectories a divergence can be seen which we believe is largely an artifact of small training datasets.} \label{fig6}
\end{figure}

At the end, one final experiment was conducted. This is one that kernel feature expectations would be unable to do and so there is no comparison to make. Using our learned reward function we replaced one of the AI players in a match to see how closely a reward generated policy would match the replaced player's trajectory. The result of this can be seen in Figure \ref{fig6}.

\section{Conclusion}\label{sec:concl}

In this paper we developed a data driven technique to define an MDP and learn a reward function which explains observed behavior. To validate this technique we generated 36 simulated engagements within a high-fidelity combat simulator and examined reward functions learned from observations of the engagement. We were able to show that by examining the reward functions learned through our technique more accurate predictions could be made about the strategy that generated the observed behavior.

In the future we plan on extending this method to a more general multi-agent formulation for IRL. Such an extension seems important for problems where many agents interact with varying goals and skill levels. Finally, we hope to provide a more sophisticated explanation for why our direct estimate method seems to beat out other RL algorithms in the limited resource use case. 


\addtolength{\textheight}{-12cm}   




\section*{ACKNOWLEDGMENT}

This work was supported by Systems Engineering, Inc and the Office of Naval Research under grant number N00014-20-C-2011.


\bibliographystyle{IEEEtran}
\bibliography{combat_irl}

\end{document}